%% file: iclr2026_conference.tex
\documentclass{article} 
\usepackage{iclr2026_conference,times}

\input{math_commands.tex}

\usepackage{hyperref}
\usepackage{url}
\usepackage{booktabs}       
\usepackage{amsfonts}       
\usepackage{xcolor}         
\usepackage{nicefrac}       
\usepackage{microtype}      
\usepackage{xcolor}         
\usepackage{amsmath} 
\usepackage{graphicx}
\usepackage{wrapfig} 
\usepackage{adjustbox}
\usepackage{multirow}
\usepackage{array}
\usepackage{colortbl}
\usepackage{caption}
\usepackage{enumitem}
\usepackage{subcaption}
\usepackage[utf8]{inputenc} 
\usepackage[T1]{fontenc}    
\usepackage{url}            
\usepackage{booktabs}       
\usepackage{amsfonts}       
\usepackage{nicefrac}       
\usepackage{microtype}      
\usepackage{amsmath} 
\usepackage{graphicx}
\usepackage{wrapfig} 
\usepackage{adjustbox}
\usepackage{multirow}
\usepackage{array}
\usepackage{colortbl}
\usepackage{caption}
\usepackage{enumitem}
\usepackage{hyperref}
\usepackage{algorithm}
\usepackage{algorithmic}
\usepackage{xcolor}
\usepackage[table]{xcolor}
\usepackage{subcaption}
\usepackage{siunitx}
\usepackage[dvipsnames]{xcolor}

\title{CausalAffect: Causal Discovery \\ for Facial Affective Understanding}

\author{%
  Guanyu Hu\textsuperscript{1,2} 
  Tangzheng Lian\textsuperscript{3} 
  Dimitrios Kollias\textsuperscript{2} 
  Oya Celiktutan\textsuperscript{3} 
  Xinyu Yang\textsuperscript{1} 
  \\
  \textsuperscript{1}Xi'an Jiaotong University 
  \textsuperscript{2}Queen Mary University of London 
  \textsuperscript{3}Kings College London 
}

\iclrfinalcopy 
\begin{document}

\maketitle

\begin{abstract}
Understanding human affect from facial behavior requires not only accurate recognition but also structured reasoning over the latent dependencies that drive muscle activations and their expressive outcomes. 
Although Action Units (AUs) have long served as the foundation of affective computing, existing approaches rarely address how to infer psychologically plausible causal relations between AUs and expressions directly from data. 
We propose \textbf{CausalAffect}, the first framework for causal graph discovery in facial affect analysis. 
CausalAffect models AU$\rightarrow$AU and AU$\rightarrow$Expression dependencies through a two-level polarity and direction aware causal hierarchy that integrates population-level regularities with sample-adaptive structures. 
A feature-level counterfactual intervention mechanism further enforces true causal effects while suppressing spurious correlations. 
Crucially, our approach requires neither jointly annotated datasets nor handcrafted causal priors, yet it recovers causal structures consistent with established psychological theories while revealing novel inhibitory and previously uncharacterized dependencies. 
Extensive experiments across six benchmarks demonstrate that CausalAffect advances the state of the art in both AU detection and expression recognition, establishing a principled connection between causal discovery and interpretable facial behavior. All trained models and source code will be released upon acceptance.

\end{abstract}

\section{Introduction}

Understanding human emotions from facial behavior is a core task in affective computing, with broad implications for human–computer interaction, assistive technologies, and robotics~\citep{picard2000affective}. 
A key bridge from facial behavior to emotion understanding is the Facial Action Coding System (FACS)~\citep{ekman1978facial}, a widely adopted framework that decomposes facial expressions into a set of Action Units (AUs), which are elemental muscle movements associated with facial activity. Capturing co-activation structures, both among AUs (i.e., AU$\rightarrow$AU interactions) and between AUs and expressions (i.e., AU$\rightarrow$Expression linkages), has been shown to significantly improve \textit{Action Unit Detection (AUD)} and \textit{Facial Expression Recognition (FER)}.
Existing approaches for modeling such dependencies fall into three paradigms: 
\textbf{(i) Cognitive prior-based methods}, which incorporate expert-defined co-activation patterns from psychological studies~\citep{du2014compound}; 
\textbf{(ii) Data-driven learning methods}, which leverage relational inductive biases (e.g., graph neural networks, GNNs) to learn dependencies from data~\citep{song2021uncertain,luo2022learning,liu2020relation}; and  
\textbf{(iii) Statistical co-occurrence approaches}, which infer relational patterns from co-activation statistics in jointly annotated datasets~\citep{song2015exploiting, kollias2021affect,kollias2024distribution}.

Although grounded in different principles, all above paradigms suffer from four key limitations:  
\textbf{(i) Lack of psychological plausibility:}  
As evidenced in Fig.~\ref{fig:au_expr_heatmap} (GNN), data-driven learning approaches often induce entangled patterns that, while superficially interpretable, diverge from human-aligned causal structures.  Similarly, statistical co-occurrence approaches only reflects dataset-specific frequencies rather than genuine causal mechanisms and are susceptible to demographic biases (e.g., culture, age, gender);
\textbf{(ii) Dependence on joint annotations:}  
Most methods rely on datasets with joint annotations, which are scarce and rarely overlapping, leaving most single-task datasets underutilized;
\textbf{(iii) Limited to rigid global relations:}  
Although cognitive studies emphasize psychological plausibility, they restricted to fixed population-level relations, overlooking context-adaptive patterns.  For example, FACS~\citep{ekman1978facial} links AU6 (cheek raiser) and AU12 (lip-corner puller) to happiness, yet both \textit{Duchenne} and \textit{Social} smiles involve these AUs but convey distinct meanings (see analysis in Section~\ref{sec:sa}); 
\textbf{(iv) Neglect of directionality and inhibitory effects:}  
All these studies often treat AU relations as symmetric, especially in AU$\rightarrow$AU interactions, whereas in practice muscle activations are directional (e.g., AU9 (nose wrinkler) mechanically triggering AU10 (upper lip raiser), but not vice versa).  
They also rarely account for inhibitory relations, where certain AUs weaken or mask emotional expressions (e.g., AU6 reducing Sadness or AU12 diminishing Disgust), thereby limiting the modeling of the full spectrum of affective dynamics.

To address these limitations, we introduce \textbf{CausalAffect}, a unified graph framework for discovering \textit{directed (asymmetric), polarity-aware (excitatory and inhibitory), and both population-level and sample-adaptive} causal relationships between AUs and expressions. Operating in a weakly supervised setting, CausalAffect requires neither co-annotated labels nor handcrafted priors. A key innovation is the \textbf{feature-level counterfactual intervention}, which enables causal discovery by perturbing disentangled AU representations. Unlike prior counterfactual methods that manipulate high-level attributes (e.g., age, gender) in image space using generative models~\citep{cheong2022counterfactual,liu2024flipping,ramesh2024synthetic}—approaches that are computationally expensive, semantically coarse, and ill-suited to structured causal reasoning—CausalAffect intervenes directly in the AU latent space, yielding semantically faithful, computationally efficient, and interpretable causal modeling.  
In summary, our contributions are threefold:
\begin{itemize}[leftmargin=1.5em, labelsep=1em, itemsep=3pt, parsep=0pt]
\item \textbf{Causal Discovery.}  
To the best of our knowledge, this is the first data-driven framework that learns asymmetric, polarity-aware causal dependencies in both AU$\rightarrow$AU and AU$\rightarrow$Expression relations, yielding structures that are causally grounded and psychologically plausible.
\item \textbf{CausalAffect Framework.} We introduce CausalAffect, a unified causal discovery framework that jointly captures population-level and sample-adaptive causal structure through global graph induction, sample-adaptive graph refinement, and feature-level counterfactual intervention.
\item \textbf{Feature-level Counterfactual Intervention.} We propose a bottleneck-based intervention mechanism that perturbs latent AU features directly, eliminating the need for image-level synthesis and its associated complexity. This design enables efficient and scalable counterfactual reasoning, while injecting causal supervision to promote structurally meaningful dependency learning.
\item \textbf{State-of-the-Art Performance.}  
Extensive experiments on six benchmarks, spanning both AU detection and expression recognition in image and video settings, demonstrate that CausalAffect achieves state-of-the-art results under diverse training configurations.
\end{itemize}


\section{Related Work}

Existing work on facial affect analysis explores both structured relation modeling and causal or counterfactual reasoning, but each direction has notable limitations. Prior methods model AU–AU and AU–expression dependencies using expert-curated FACS rules \citep{ekman1978facial,du2014compound} or data-driven graph and attention architectures \citep{song2021dynamic, wang2023spatial, liu2020relation}, yet these approaches often produce dense, undirected affinity graphs that lack directionality, polarity, and human-aligned interpretability \citep{kakkad2023survey, wang2023graph}. Co-occurrence–based relations \citep{eleyan2009co, kollias2021affect} suffer from dataset bias \citep{chen2021understanding, dominguez2024metrics} and require joint AU–expression labels. Meanwhile, causal and counterfactual models in computer vision \citep{cheong2022counterfactual,chen2022towards,li2024counterfactual,pan2024counterfactual} predominantly intervene in pixel-level or coarse semantic spaces, often relying on generative models \citep{melistas2024benchmarking,ramesh2024synthetic} and assuming fixed global causal structures \citep{gao2021causal}. None of these methods address the need for directed, polarity-aware, and sample-adaptive causal reasoning. Our work differs by performing causal discovery, enabling weakly supervised multi-dataset learning, and introducing efficient feature-level counterfactual interventions in a disentangled AU latent space.

\section{Problem Formulation}

The objective is to uncover \emph{causally grounded} and \emph{psychologically plausible} structures that govern dependencies among AUs and between AUs and expressions, without relying on co-occurring label annotations. We assume access to two  type of disjoint heterogenious datasets: AU-labeled dataset $\mathcal{D}_{\text{AU}}=\{(I_k^{\text{AU}},y_k^{\text{AU}})\}_{k=1}^{K_{\text{AU}}}$ with multi-label annotations $y_k^{\text{AU}}\in\{0,1\}^{N_{\text{AU}}}$, and expression-labeled dataset $\mathcal{D}_{\text{Expr}}=\{(I_k^{\text{Expr}},y_k^{\text{Expr}})\}_{k=1}^{K_{\text{Expr}}}$ with categorical labels $y_k^{\text{Expr}}\in\{1,\dots,N_{\text{Expr}}\}$. Each image $I$ is encoded into a set of disentangled AU-level embeddings $\mathbf{F}_{\text{AU}}=\{\mathbf{f}_{\text{AU}}^{(i)}\}_{i=1}^{N_{\text{AU}}}$, where $\mathbf{f}_{\text{AU}}^{(i)}\in\mathbb{R}^d$ denotes the representation of the $i$-th AU. We formalize causal discovery as learning two types of directed graphs: \textbf{AU$\rightarrow$AU} graphs, where target nodes are causally aggregated AU embeddings capturing inter-AU dependencies, and \textbf{AU$\rightarrow$Expression} graphs, where target nodes are expression representations causally aggregated from AU features. For each type, we maintain two levels of graphs: a \textbf{global causal graph} $\mathcal{G}_g$ that models population-level dependencies consistent with cognitive and psychological literature (ensuring psychological plausibility), and a \textbf{sample-adaptive causal graph} $\mathcal{G}_s(I)$ that refines the global relations to capture instance-specific causal patterns. Depending on the dataset, the framework can be instantiated for AU-only graph discovery (AU detection), AU$\rightarrow$Expression graph discovery (expression recognition), or jointly in a multi-task setting.

\begin{figure}[t]
    \centering
    \includegraphics[width=1\textwidth]{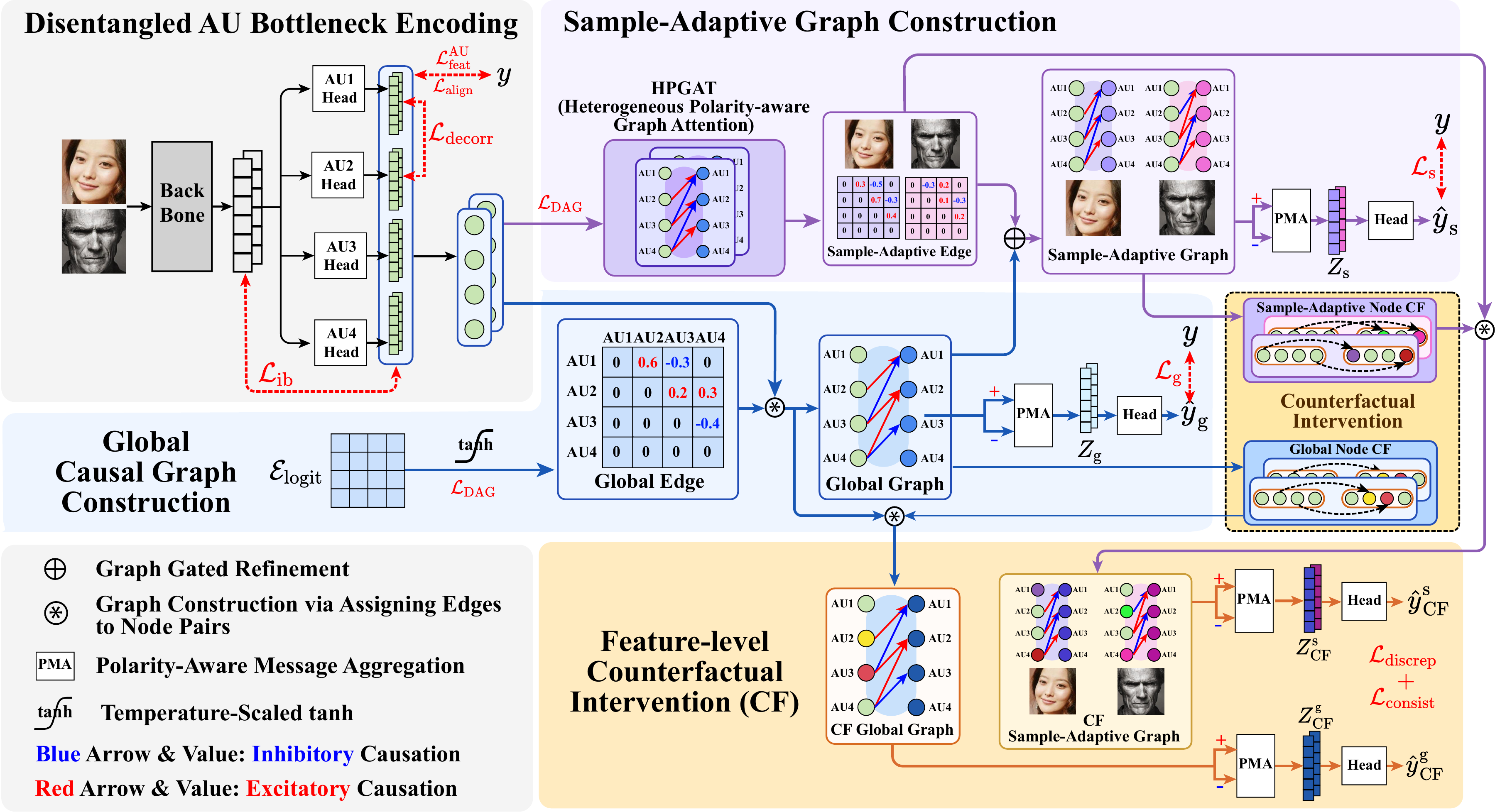}
      \captionsetup{font=footnotesize} 
\caption{Overview of \textbf{CausalAffect}, consisting of four key components: (1) \textbf{Disentangled AU Bottleneck Encoding}, which extracts semantically independent AU features; (2) \textbf{Global Causal Graph Construction}, which learns a polarity-aware acyclic graph capturing population-level dependencies; (3) \textbf{Sample-Adaptive Graph Refinement}, which adjusts the global graph for each instance via residual adaptation; and (4) \textbf{Feature-Level Counterfactual Intervention}, which enforces causal inductive bias through latent perturbations and factual–counterfactual consistency. The figure illustrates the AU$\rightarrow$AU branch; the AU$\rightarrow$Expression branch is analogous and omitted for brevity.}
    \label{fig:model}
\end{figure}

\section{CausalAffect Framework}
In this section, we present the  \textbf{CausalAffect} framework, the  overall structure is illustrated in Fig.~\ref{fig:model}.

\subsection{Disentangled AU Bottleneck Encoding}

The first stage of CausalAffect focuses on learning semantically meaningful and explicitly disentangled AU features, which form the foundation for reliable causal discovery. Disentanglement is essential because each AU representation must isolate the information specific to that AU, without identity cues, shared nuisance factors, or spurious co-variations. By ensuring that AU features encode only their intended semantic content, this stage removes confounders, suppresses correlated noise, and enables both the construction of meaningful AU–AU and AU–Expression causal structures and the execution of effective feature-level counterfactual interventions.

To achieve this, we first obtain global visual features from the input image $I$ and then project them into AU-specific representations.  
Our framework is backbone-agnostic, we adopt ConvNeXt followed by a Squeeze-and-Excitation (SE) block~\citep{hu2018squeeze} to produce a compact embedding $z_{\text{img}} \in \mathbb{R}^D$.  
On top of the embedding, we introduce $N_{\text{AU}}$ lightweight projection heads $\{\phi_i\}_{i=1}^{N_{\text{AU}}}$, where each $\phi_i: \mathbb{R}^D \rightarrow \mathbb{R}^d$ maps $z_{\text{img}}$ into an AU-specific feature $\mathbf{f}_{\text{AU}}^{(i)}$.  
This yields a feature matrix  
$\mathbf{F}_{\text{AU}} = [\phi_1(z_{\text{img}}), \ldots, \phi_{N_{\text{AU}}}(z_{\text{img}})]^\top \in \mathbb{R}^{N_{\text{AU}} \times d}$.
To ensure that each AU representation is discriminative.  
Each $\mathbf{f}_{\text{AU}}^{(i)}$ is further fed into a binary classifier to predict the activation $\hat{y}_{\text{AU}}^{(i)}$, supervised by the binary cross-entropy (BCE) loss:  
$\mathcal{L}_{\text{feat}}^{\text{AU}} = \frac{1}{N_{\text{AU}}} \sum_{i=1}^{N_{\text{AU}}} \text{BCE}(\hat{y}_{\text{AU}}^{(i)},\, y_{\text{AU}}^{(i)})$.

\textbf{Information Bottleneck via HSIC-Based Disentanglement.}
To ensure that each AU-specific representation preserves only task-relevant information while discarding confounding factors (e.g., age, gender, identity) embedded in $z_{\text{img}}$, we adopt the Deep Information Bottleneck principle~\citep{tishby2000information,tishby2015deep} for disentanglement. This principle imposes two core constraints:
\textbf{(i) \emph{Minimize $\mathcal{L}_{\text{ib}}$}:} the statistical dependence between AU representation $\mathbf{f}_{\text{AU}}^{(i)}$ and $z_{\text{img}}$ to suppress task-irrelevant signals;  
\textbf{(ii) \emph{Maximize $\mathcal{L}_{\text{align}}$}:} the dependence between $\mathbf{f}_{\text{AU}}^{(i)}$ and its corresponding label to retain discriminative semantics. 
Beyond these, we introduce an additional objective:  \textbf{(iii) \emph{Minimize $\mathcal{L}_{\text{decorr}}$}:} the statistical dependence between different AU representations $\small{\mathbf{f}_{\text{AU}}^{(i)}}$ and $\small{\mathbf{f}_{\text{AU}}^{(j)}}$ $(i \ne j)$ to reduce representational redundancy. This constraint promoting semantically disentangled and structurally independent features, which is essential for reliable causal attribution. Since direct mutual information estimation is computationally unstable and difficult to optimize, we instead adopt the HSIC~\citep{gretton2005measuring} as a reliable, non-parametric surrogate for all dependency objectives. HSIC enables efficient and theoretically grounded disentanglement through kernel-based statistical measurement. The three objectives defined as follows (see \textbf{Appendix E} for HSIC computation):
{\small
\begin{equation}
    \mathcal{L}_{\text{ib}} = \frac{1}{N_{\text{AU}}} \sum_{i} \mathcal{H}(\mathbf{f}_{\text{AU}}^{(i)}, z_{\text{img}}),\quad
    \mathcal{L}_{\text{align}} = -\frac{1}{N_{\text{AU}}} \sum_{i} \mathcal{H}(\mathbf{f}_{\text{AU}}^{(i)}, \tilde{y}_{\text{AU}}^{(i)}),\quad
    \mathcal{L}_{\text{decorr}} = \frac{1}{N_{\text{AU}}^2} \sum_{i \ne j} \mathcal{H}(\mathbf{f}_{\text{AU}}^{(i)}, \mathbf{f}_{\text{AU}}^{(j)}).
\end{equation}}\noindent
where  $\mathcal{H}(\cdot, \cdot)$ denotes the HSIC dependence measure, $N_{\text{AU}}$ denotes the number of AUs; and $\tilde{y}_{\text{AU}}^{(i)}$ represents the (pseudo-)label for AU $i$, which uses ground-truth labels when available and model predictions otherwise. Higher HSIC values  indicate stronger dependence. 

\textbf{Total AU Bottleneck Loss.}
The final objective for learning disentangled AU bottleneck features is:
\begin{equation}
    \mathcal{L}_{\text{AU}} = \mathcal{L}_{\text{feat}}^{\text{AU}} + \lambda_{\text{ib}} \cdot \mathcal{L}_{\text{ib}} + \lambda_{\text{decorr}} \cdot \mathcal{L}_{\text{decorr}} + \lambda_{\text{align}} \cdot \mathcal{L}_{\text{align}},
\end{equation}
where $\lambda_{\text{ib}}, \lambda_{\text{decorr}}, \lambda_{\text{align}}$ control the strength of each regularization term.

\subsection{Global Causal Graph Construction (GC Graph)}

The second stage of CausalAffect aims to capture stable, population-level causal dependencies for both \textit{AU$\rightarrow$AU} and \textit{AU$\rightarrow$Expression} interactions. This is achieved by learning a shared adjacency matrix whose edges are directed (asymmetric) and polarity-aware, two properties that are essential for reflecting the true causal nature of facial behavior. AU$\rightarrow$Expression relations are inherently \textbf{unidirectional and DAG-structured}, as expressions arise \emph{as downstream outcomes} of underlying AU activations. In contrast, AU$\rightarrow$AU relations often exhibit \textbf{positive reciprocal causation}, where one muscle unit reinforces or stabilizes the activation of another; therefore, we incorporate a soft-DAG constraint as a \emph{regularization mechanism} that encourages the learned AU$\rightarrow$AU structure to remain \textbf{sparse, directional, and dominated by strong causal influences}. Polarity-aware edges are equally crucial, since facial musculature expresses both \textbf{excitatory} (co-activating) and \textbf{inhibitory} (suppressing or dampening) effects.

Formally, we define the global graph as $\mathcal{G}_\text{g} = (\mathcal{N}_\text{g, source}, \mathcal{N}_\text{g, target}, \mathcal{E}_\text{g})$, where the source nodes $\mathcal{N}_\text{g, source}$ are the disentangled AU representations $\mathbf{F}_{\text{AU}}$ shared across both graph types. The target nodes $\mathcal{N}_\text{g, target}$ correspond to representations causally aggregated from $\mathcal{N}_\text{g, source}$ (expressions in AU$\rightarrow$Expression, AUs in AU$\rightarrow$AU). The edge weights $\mathcal{E}_\text{g} \in (-1,1)$ encode both the polarity and magnitude of causal influence, thereby specifying the directed topology from source to target. To parameterize these edges, we introduce a learnable matrix $\mathcal{E}_{\text{logit}} \in \mathbb{R}^{N_{\text{target}} \times N_{\text{source}}}$, where each entry $\mathcal{E}_{\text{logit}}^{(j,i)}$ denotes the raw, unbounded causal strength from source node $i$ to target node $j$, with $N_{\text{g, source}}$ and $N_{\text{g, target}}$ is the number of source and target nodes.
To obtain bounded edge matrix, we apply a temperature-scaled hyperbolic tangent transformation:
\begin{equation}
\mathcal{E}_{\text{g}}^{(j, i)} = \tanh\left(\tau \cdot \mathcal{E}_{\text{logit}}^{(j, i)}\right), \quad \tau = \exp(T_{\text{g}}),
\label{eq:globaledge}
\end{equation}
where $T_{\text{g}}$ is a learnable scalar temperature. This maps the logits into the interval $(-1, 1)$, producing a signed edge matrix $\mathcal{E}_{\text{g}} \in \mathbb{R}^{N_{\text{target}} \times N_{\text{source}}}$. The sign of each entry indicates whether the influence is excitatory ($>0$) or inhibitory ($<0$), while the magnitude encodes causal strength. 

\textbf{Polarity-Aware Message Aggregation.} To disentangle polarity effects, we decompose the signed edge into excitatory and inhibitory components:
\begin{equation}
\mathcal{E}_{\text{g}}^{+} = \mathbb{I}[\mathcal{E}_{\text{g}} > 0] \odot \mathcal{E}_{\text{g}}, \quad
\mathcal{E}_{\text{g}}^{-} = \mathbb{I}[\mathcal{E}_{\text{g}} < 0] \odot (-\mathcal{E}_{\text{g}}),
\label{eq:twopath}
\end{equation}
where $\mathbb{I}[\cdot]$ is the element-wise indicator function and $\odot$ denotes the Hadamard product.
To aggregate source AU nodes into target nodes (AU or Expression) through graph, we first project the disentangled AU features into two value spaces, i.e., $V^{+}_{g}=\mathbf{F}_{\text{AU}}W_{g}^{+}$ and $V^{-}_{g}=\mathbf{F}_{\text{AU}}W_{g}^{-}$, where $W_{g}^{+},W_{g}^{-}\in\mathbb{R}^{d\times d_{\text{g}}}$ are the positive and negative projection matrices, $d_{\text{g}}$ denote the dimensionality of the projected representation.
The aggregated representation is computed as $Z_{\text{g}}=\mathcal{E}_{\text{g}}^{+}V_{g}^{+}+\mathcal{E}_{\text{g}}^{-}V_{g}^{-}\in\mathbb{R}^{N_{\text{target}}\times d_{\text{g}}}$.

\textbf{AU$\rightarrow$AU Homogeneous Soft DAG Constraint.}  
From a psychological and muscular perspective, facial AUs are driven by underlying muscle activations that follow directional and structured pathways rather than arbitrary feedback loops. Unlike the AU$\rightarrow$Expression graph, which is heterogeneous and inherently acyclic due to its one-way causal semantics, the AU$\rightarrow$AU graph is homogeneous and permits self-contained interactions among AUs, making it especially prone to cycles during learning. To mitigate this, we introduce a differentiable DAG constraint to regularize the AU$\rightarrow$AU graph. Instead of enforcing strict acyclicity, we implement it as a \textbf{soft} regularization, guiding the model toward interpretable, directional dependencies while preserving flexibility to capture reciprocal or context-dependent AU interactions.
Specifically, for the learned AU$\rightarrow$AU edge matrix $\mathcal{E}_{g}^{\mathrm{AU}} \in \mathbb{R}^{N_{\mathrm{AU}} \times N_{\mathrm{AU}}}$, we enforce the constraint using the trace (denoted $\operatorname{tr}$) of its matrix exponential:
\begin{equation}
    \mathcal{L}_{\text{DAG}} = \operatorname{tr}\left(\exp\left(\mathcal{E}_{\text{g}}^{\text{AU}} \odot \mathcal{E}_{\text{g}}^{\text{AU}}\right)\right) - N_{\text{AU}},
\end{equation}
where $\odot$ is the Hadamard product. This DAG constraint not only to the global AU$\rightarrow$AU graph, but also to the sample-adaptive AU$\rightarrow$AU graphs introduced in the next section.

\textbf{Global Graph Loss.}
For AU$\rightarrow$AU graph, the prediction is $\hat{y}_{\text{AU}}=\Phi_{\text{AU}}(Z_{\text{g}}^{\text{AU}})$; 
for AU$\rightarrow$Expression graph, the prediction is $\hat{y}_{\text{Expr}}=\Phi_{\text{Expr}}(Z_{\text{g}}^{\text{Expr}})$, 
where $Z_{\text{g}}^{\text{AU}}\!\in\!\mathbb{R}^{N_{\text{AU}}\times d_{\text{g}}}$ and $Z_{\text{g}}^{\text{Expr}}\!\in\!\mathbb{R}^{N_{\text{Expr}}\times d_{\text{g}}}$ are the causally aggregated features, and $\Phi_{\text{AU}}$, $\Phi_{\text{Expr}}$ denote task-specific classifier heads.
The graph learning is supervised using the corresponding ground-truth labels $y^{\text{AU}}$ and $y^{\text{Expr}}$ via the $\mathcal{L}_{\text{g}}$ loss defined as:
\begin{equation}
    \mathcal{L}_{\text{g}} =
    \begin{cases}
        \mathcal{L}_{\text{g}}^{\text{AU}} = \text{BCE}(\hat{y}_{\text{g}}^{\text{AU}}, y^{\text{AU}}) + \lambda_{\text{DAG}}^{\text{g}} \cdot \mathcal{L}_{\text{DAG}}^{\text{g}}, & \text{AU$\rightarrow$AU Graph}, \\
        \mathcal{L}_{\text{g}}^{\text{Expr}} = \text{CE}(\hat{y}_{\text{g}}^{\text{Expr}}, y^{\text{Expr}}), & \text{AU$\rightarrow$Expression Graph},
    \end{cases}
\end{equation}
where $\text{BCE}(\cdot)$/$\text{CE}(\cdot)$ denote binary/categorical cross-entropy, $\lambda_{\text{DAG}}$ controls acyclicity constraint.

\subsection{Sample-Adaptive Causal Graph Construction (SAC Graph)}

While the global causal graph captures population-level patterns, it may overlook context-specific dependencies (e.g., Eastern vs. Western display norms, or cases where \textit{Social} and \textit{Duchenne} smiles share AU patterns but stem from different causal pathways). Cross-cultural studies by Jack \citep{jack2012facial,jack2014dynamic} show that facial expressions combine universal AU configurations with culture- and individual-specific variations. For instance, the core AU12+AU6 smile pattern is shared globally, yet Western smiles often add AUs such as AU7, AU25, and AU26, whereas East Asian smiles modulate intensity mainly through the mouth region. A single fixed graph cannot represent such variability, motivating a model that integrates both global and sample-adaptive causal structures.

To address this, we introduce sample-adaptive causal graph $\mathcal{G}_\text{s}^{(k)}=(\mathcal{N}_\text{s,source}^{(k)},\mathcal{N}_\text{s,target}^{(k)},\mathcal{E}_\text{s}^{(k)})$, dynamically constructed for each input $I^{(k)}$. This graph refines the global structure via an adaptive edge residual, allowing the model to capture context-dependent variations.  
To instantiate $\mathcal{G}_\text{s}^{(k)}$, we design a multi-layer attention module termed \textit{Heterogeneous Polarity-aware Graph Attention (HPGAT)}. Standard attention-based graph models such as GAT and GATv2~\citep{brody2021attentive} cannot model the inherently \textbf{heterogeneous} and \textbf{polarity-dependent} causal mechanisms in AU$\rightarrow$AU and AU$\rightarrow$Expression reasoning, since they assume homogeneous node types and restrict attention weights to be non-negative, preventing the representation of inhibitory effects. HPGAT introduces three key advances:  
\textbf{(i) Heterogeneous node modeling}, enabling cross-type AU$\rightarrow$Expr aggregation;  
\textbf{(ii) Polarity-aware edge learning}, capturing both excitatory and inhibitory causal effects within $(-1,1)$;  
\textbf{(iii) Gated causal residual integration}, adaptively refining the global causal structure at the instance level.  
Together, these capabilities allow HPGAT to learn \textbf{sample-adaptive, polarity-aware causal graphs} that cannot be expressed by conventional homogeneous attention mechanisms.

For each sample $k$, HPGAT infers cross-node attentional affinities between $\mathcal{N}_\text{s, target}^{(k)}$ and $\mathcal{N}_\text{s, source}^{(k)}$. At the final layer, the learned edge attention is incorporated as a residual refinement to the global edge matrix $\mathcal{E}_{\text{g}}$ in Eq.~\ref{eq:globaledge}, yielding the final sample-specific graph $\mathcal{E}_{\text{s}}^{(k)}$.
Concretely, for both the AU$\rightarrow$AU and AU$\rightarrow$Expression graphs, source AU nodes $\mathcal{N}_\text{s, source}^{(k)}$ serve as attention \textbf{Keys} and \textbf{Values} and are fixed across layers, i.e., $K_s^{(k, l)}=V_s^{(k, l)}=\mathcal{N}_\text{s, source}^{(k)}=\mathbf{F}^{(k)}_\text{AU}$. 
The \textbf{Queries} depend on the target node type $\mathcal{N}^{(k)}_\text{s, target}$. At the first layer ($l=1$), for the AU$\rightarrow$AU graph we set $Q^{(1)}=\mathbf{F}^{(k)}_\text{AU}$, whereas for the AU$\rightarrow$Expression graph we initialize queries with learnable expression prototypes $\mathbf{P}_\text{Expr}\in\mathbb{R}^{N_\text{Expr}\times d}$. For subsequent layers ($l>1$), queries are updated recursively as $Q^{(k, l)}=\mathcal{N}_\text{s, target}^{(l-1)}$.
At layer $l$, the HP-GAT computes the signed polarity-aware edge matrix $\mathcal{E}_s^{(k,l)} \in \mathbb{R}^{N_{\text{target}} \times N_{\text{source}}}$ as:
{\small
\begin{equation}
\mathcal{E}_s^{(k,l)} = 
\tanh\!\Big(
    \mathbf{a}^{\top} \, \sigma\!\big( 
        W^{(k,l)}_{Q} Q^{(k,l)} 
        + W^{(k,l)}_{K} K_{s}^{(k,l)} 
    \big)
\Big),
\end{equation}
}
\noindent
where $\sigma(\cdot)$ denotes the LeakyReLU activation, $W^{(k,l)}_{Q}, W^{(k,l)}_{K} \in \mathbb{R}^{d \times d_h}$ are learnable projection matrices, $\mathbf{a} \in \mathbb{R}^{d_h}$ is a learnable attention vector, and $d_h$ denotes the hidden dimension of the projected feature space. The $\tanh(\cdot)$ bounds edge weights to $(-1,1)$.

\textbf{Polarity-Aware Message Aggregation.}  
Analogous to Eq.~\ref{eq:twopath}, decompose causal polarity components:
\begin{equation}
\label{eq:sa_edge}
    \mathcal{E}_{s}^{+(k,l)} = \mathbb{I}\!\left[\mathcal{E}_s^{(k,l)} > 0\right] \odot \mathcal{E}_s^{(k,l)}, 
    \quad
    \mathcal{E}_{s}^{-(k,l)} = \mathbb{I}\!\left[\mathcal{E}_s^{(k,l)} < 0\right] \odot \big(-\mathcal{E}_s^{(k,l)}\big).
\end{equation}
The updated target node embedding $\mathcal{N}^{(k,l)}_{\text{s,target}} = Z_s^{(k,l)}$ are obtained by aggregating the polarity-aware edges over the value projections: $Z_s^{(k,l)} = \mathcal{E}_{s}^{+(k,l)} V_s^{(k,l)} W^{+(k,l)}_{V} + \mathcal{E}_{s}^{-(k,l)} V_s^{(k,l)} W^{-(k,l)}_{V}$, where $W^{+(k,l)}_{V}, W^{-(k,l)}_{V} \in \mathbb{R}^{d \times d_h}$  are learnable projections for excitatory and inhibitory messages.

\textbf{Sample-Adaptive Graph Gated Refinement.}
At the final layer $L$, we refine the global edge structure $\mathcal{E}_\text{g}$ with the sample-specific residual edge $\mathcal{E}^{(k,L)}$ through a learnable gating mechanism. 
In particular, a non-linear projection layer $\mathcal{G}(\cdot)$ maps $\mathcal{E}_s^{(k,L)}$ into a scalar gating value, and fused edge computed as:
\begin{equation}
    \mathcal{E}_\text{s}^{(k)} = 
    \sigma\!\left(\mathcal{G}\!\big(\mathcal{E}_s^{(k,L)}\big)\right) \odot \mathcal{E}_\text{g}
    + \Big(1 - \sigma\!\left(\mathcal{G}\!\big(\mathcal{E}_s^{(k,L)}\big)\right)\Big) \odot \mathcal{E}_s^{(k,L)},
\end{equation}
where $\sigma(\cdot)$ denotes the sigmoid function. 
Following Eq.~\ref{eq:sa_edge}, the edge matrix is further decomposed into excitatory and inhibitory components, $\mathcal{E}_{s}^{+(k,L)}$ and $\mathcal{E}_{s}^{-(k,L)}$, and the final sample-adaptive target node are updated as
$\mathcal{N}^{(k)}_{\text{s, target}} = Z_s^{(k,L)} = \mathcal{E}_{s}^{+(k,L)} V_s^{(k,L)} W^{+(k,L)}_{V} 
+ \mathcal{E}_{s}^{-(k,L)} V_s^{(k,L)} W^{-(k,L)}_{V}$.

\textbf{Sample-Adaptive Graph Loss.}
For the AU$\rightarrow$AU graph, the prediction is 
$\hat{y}_{\text{s}}^{\text{AU}} = \Phi_{\text{AU}}(Z_{\text{s}}^{\text{AU}})$; 
for the AU$\rightarrow$Expression branch, the prediction is 
$\hat{y}_{\text{s}}^{\text{Expr}} = \Phi_{\text{Expr}}(Z_{\text{s}}^{\text{Expr}})$, 
$\Phi_{\text{AU}}$, $\Phi_{\text{Expr}}$ denote task-specific classifier heads.  
The learning is supervised using the labels $y^{\text{AU}}$ and $y^{\text{Expr}}$ via the $\mathcal{L}_{\text{s}}$ loss defined as:
\begin{equation}
    \mathcal{L}_{\text{s}} =
    \begin{cases}
        \mathcal{L}_{\text{s}}^{\text{AU}} = \text{BCE}(\hat{y}_{\text{s}}^{\text{AU}}, y^{\text{AU}}) + \lambda_{\text{DAG}}^{\text{s}} \cdot \mathcal{L}_{\text{DAG}}^{\text{s}}, & \text{AU$\rightarrow$AU Graph}, \\
        \mathcal{L}_{\text{s}}^{\text{Expr}} = \text{CE}(\hat{y}_{\text{s}}^{\text{Expr}}, y^{\text{Expr}}), & \text{AU$\rightarrow$Expression Graph},
    \end{cases}
\end{equation}
where $\lambda_{\text{DAG}}^{\text{s}}$ controls the acyclicity regularization applied only to the AU$\rightarrow$AU graph.

\subsection{Feature-level Counterfactual Intervention}

While the global and sample-adaptive graphs capture structural dependencies, they may still encode spurious correlations. To enforce relationships that are truly \textit{causal} rather than merely \textit{correlational}, we introduce a \textbf{Feature-level Counterfactual Intervention (CF)} module. The core principle is that if a source node genuinely exerts a causal influence on a target, then intervening on the source should induce a meaningful change in the target’s representation and prediction, whereas intervening on non-causal nodes should produce little to no effect.

We realize this via a thresholded soft mask with edge-aware saliency, 
which perturbs source node embeddings conditioned on the graph. 
By contrasting model behaviors under factual and counterfactual conditions, the model receives consistency and discrepancy signals that guide graph refinement, reinforcing informative edges and suppressing spurious ones.
Specifically, given a graph edge matrix 
$\mathcal{E} \in (-1,1)^{N_{\text{target}} \times N_{\text{source}}}$ 
(from either global or sample-adaptive graphs), 
we compute a soft importance mask 
$\mathbf{M}_{\text{soft}} \in (0,1)^{N_{\text{target}} \times N_{\text{source}}}$ as
\begin{equation}
\mathbf{M}_{\text{soft}} = \sigma\!\left( \gamma \cdot (|\mathcal{E}| - \boldsymbol{\theta}) \right),
\end{equation}
where $\boldsymbol{\theta} \in [0,1]^{N_{\text{target}}}$ is a learnable threshold vector,
$\gamma$ is a sharpness factor, and $\sigma(\cdot)$ is the element-wise sigmoid. 
Larger $\gamma$ yields sharper distinctions between relevant and irrelevant source nodes.
For each target node $j$, we extract a row vector 
$\mathbf{m}^{(j)} \in [0,1]^{N_{\text{source}}}$, 
and define two complementary intervention masks. 
The \textbf{consistency mask} is given by 
$\smash{\mathbf{m}_{\text{consist}}^{(j)}} = 1 - \mathbf{m}^{(j)}$, which masks non-causal source nodes, 
while the \textbf{discrepancy mask} is defined as 
$\smash{\mathbf{m}_{\text{discrep}}^{(j)}} = \mathbf{m}^{(j)}$, which masks causal nodes.
Counterfactual interventions are implemented by injecting Gaussian perturbations:
\begin{equation}
\mathbf{F}_{\text{AU-CF}}^{(j,\star)} = \mathbf{F}_{\text{AU}} + \boldsymbol{\epsilon}^{(j)} \odot \mathbf{m}_{\star}^{(j)}, 
\quad \boldsymbol{\epsilon}^{(j)} \sim \mathcal{N}(0, \tau^2), 
\quad \star \in \{\text{consist}, \text{discrep}\}.
\end{equation}
executing graph aggregation yields counterfactual target node representations $Z_{\text{CF}}^{(j,\star)}$ and predictions $\hat{y}_{\text{CF}}^{(j,\star)}$,
$\tau$ controls the perturbation. We define two complementary losses to guide causal behavior:

\textbf{Consistency loss} $\mathcal{L}_{\text{consist}}$: Guarantees output invariance under non-causal source perturbations:
{\small
\begin{equation}
\mathcal{L}_{\text{consist}} = \frac{1}{N_{\text{target}}} \sum_{j=1}^{N_{\text{target}}} \bigl[
\delta_{\text{feat}} \cdot \underbrace{(1 - \cos(Z^{(j)}, Z_{\text{CF}}^{(j,\text{consist})}))}_{\text{feature consistency}} +
\delta_{\text{logit}} \cdot \underbrace{\mathcal{D}^{\text{logit}}_{\text{consist}}(\hat{y}^{(j)}, \hat{y}_{\text{CF}}^{(j,\text{consist})})}_{\text{logit consistency}}
\bigr],
\end{equation}}\noindent 
\textbf{Discrepancy loss} $\mathcal{L}_{\text{discrep}}$: Ensures interventions on causal source nodes induce meaningful changes:
{\small
\begin{equation}
\mathcal{L}_{\text{discrep}} = \frac{1}{N_{\text{target}}} \sum_{j=1}^{N_{\text{target}}} \bigl[
\eta_{\text{feat}} \cdot \underbrace{\left(1 + \cos(Z^{(j)}, Z_{\text{CF}}^{(j,\text{discrep})})\right)}_{\text{feature discrepancy}} +
\eta_{\text{logit}} \cdot \underbrace{\left(1 - \mathcal{D}^{\text{logit}}_{\text{discrep}}(\hat{y}^{(j)}, \hat{y}_{\text{CF}}^{(j,\text{discrep})})\right)}_{\text{logit discrepancy}}
\bigr],
\end{equation}
}\noindent 
where $\delta_{\text{feat}}, \eta_{\text{feat}}$ and $\delta_{\text{logit}}, \eta_{\text{logit}}$ are the feature- and logit-level weights, $\cos(\cdot,\cdot)$ denotes cosine similarity, and $\mathcal{D}^{\text{logit}}_{\text{consist}}(\cdot,\cdot)$ is a logit-level discrepancy (MSE for AU, KL divergence for expression). 

The counterfactual intervention loss, weighted by $\lambda_{\text{consist}}$ and $\lambda_{\text{discrep}}$, is defined as:
\begin{equation}
\mathcal{L}_{\text{cf}} = \lambda_{\text{consist}} \cdot \mathcal{L}_{\text{consist}} + \lambda_{\text{discrep}} \cdot \mathcal{L}_{\text{discrep}}.
\end{equation}

\subsection{Unified Training Objective}

CausalAffect is trained end-to-end under a weakly supervised setting, where AU and expression labels are disjoint across samples. To accommodate this, we apply prediction-related losses ($\mathcal{L}_{\text{feat}}^{\text{AU}}, \mathcal{L}^{\text{g}}_{\text{AU}}, \mathcal{L}^{\text{s}}_{\text{AU}}, \mathcal{L}^{\text{g}}_{\text{Expr}}, \mathcal{L}^{\text{s}}_{\text{Expr}}$) only to samples that contain the respective ground-truth annotations. In contrast, the representation and regularization objectives ($\mathcal{L}_{\text{ib}}, \mathcal{L}_{\text{decorr}}, \mathcal{L}_{\text{align}}, \mathcal{L}_{\text{cf}}$) are applied to all samples, regardless of label availability. This design enables \textbf{weakly supervised joint learning} across disjoint datasets.
We define the total objective for each branch as follows:
\begin{equation}
    \mathcal{L}_{\text{total}}^{\text{AU}} = \mathcal{L}_{\text{AU}} + \mathcal{L}^{\text{g}}_{\text{AU}} + \mathcal{L}^{\text{s}}_{\text{AU}} + \mathcal{L}_{\text{cf}}, \qquad
    \mathcal{L}_{\text{total}}^{\text{Expr}} = \mathcal{L}^{\text{g}}_{\text{Expr}} + \mathcal{L}^{\text{s}}_{\text{Expr}} + \mathcal{L}_{\text{cf}}.
\end{equation}
The final training loss combines both branches:
    $\mathcal{L} = \mathcal{L}_{\text{total}}^{\text{AU}} + \lambda_{\text{Expr}} \cdot \mathcal{L}_{\text{total}}^{\text{Expr}}$,
where $\lambda_{\text{Expr}}$ controls the relative weighting of the expression modeling task.

\section{Experiments}

To demonstrate that CausalAffect generalizes from \textbf{static images} to \textbf{dynamic videos}, we evaluate on six datasets. For AU detection, we use three video datasets (BP4D, DISFA, GFT) and one large-scale in-the-wild image dataset (EmotioNet). For expression recognition, since video analysis primarily relies on identifying emotions from key frames, we instead adopt two large-scale in-the-wild image datasets (RAF-DB, AffectNet), providing greater diversity in environment, identity, and illumination.

Additional details are provided in the supplementary material, including 
\textbf{Appendix A:} Detailed Ablation Analysis, 
\textbf{Appendix B:} Sensitivity to AU Composition, 
\textbf{Appendix C:} Detailed Global Causal Relation Analysis, 
\textbf{Appendix D:} Extensive Sample-Adaptive Case Studies, 
\textbf{Appendix E:} HSIC computation, 
\textbf{Appendix F:} Implementation Details, Sensitivity Analysis of Regularization Parameters, 
and \textbf{Appendix G:} Training and Inference Efficiency.

\begin{table*}[ht]
\centering
\renewcommand{\arraystretch}{1.2}
\caption{
{\small
Comparison of CausalAffect with SOTA methods. 
Results are reported in F1-score (\%); For RAF-DB average accuracy (\%) is used. 
\textbf{SG} denotes single-dataset training; 
\textbf{+BP4D/DISFA/GFT/EmotioNet/RAF-DB} indicates joint training with the corresponding dataset; 
\textbf{+All} denotes training with all datasets. Statistical significance is evaluated using paired t-tests over five matched runs, CausalAffect achieves a significant improvement with $p = 0.03$ ($p<0.05$ are considered statistically significant).}
}
\begin{subtable}{\textwidth}
\centering
\renewcommand{\arraystretch}{1.2}
\normalsize
\begin{adjustbox}{max width=1\textwidth}
\begin{tabular}{lccccccccccccc||ccccccccc}
\toprule
\multirow{2}{*}{\textbf{Methods}} & \multicolumn{13}{c||}{\textbf{BP4D (AU, Video DB)}} & \multicolumn{9}{c}{\textbf{DISFA (AU, Video DB)}} \\
\cmidrule(lr){2-14} \cmidrule(lr){15-23}
& \textbf{1} & \textbf{2} & \textbf{4} & \textbf{6} & \textbf{7} & \textbf{10} & \textbf{12} & \textbf{14} & \textbf{15} & \textbf{17} & \textbf{23} & \textbf{24} & \textbf{Avg.} 
& \textbf{1} & \textbf{2} & \textbf{4} & \textbf{6} & \textbf{9} & \textbf{12} & \textbf{25} & \textbf{26} & \textbf{Avg.} \\
\midrule
\textbf{JÂA-Net~\citep{jaa2021}}                   & 53.8             & 47.8             & 58.2             & 78.5             & 75.8             & 82.7             & 88.2             & 63.7             & 43.3             & 61.8             & 45.6             & 49.9             & 62.4             & 62.4             & 60.7             & 67.1             & 41.1             & 45.1             & 73.5             & 90.9             & 67.4             & 63.5 \\
\textbf{PIAP~\citep{piap}}                         & 54.2             & 47.1             & 54.0             & 79.0             & 78.2             & \textbf{86.3}    & \textbf{89.5}    & 66.1             & 49.7             & 63.2             & 49.3             & 52.0             & 64.1             & 50.2             & 51.8             & 71.9             & 50.6             & 54.5             & \textit{79.7}    & \underline{94.1} & 57.2             & 63.8 \\
\textbf{FAUT~\citep{faut}}                         & 51.7             & \textit{49.3}    & 61.0             & 77.8             & 79.5             & 82.9             & 86.3             & 67.6             & 51.9             & 63.0             & 43.7             & 56.3             & 64.2             & 46.1             & 48.6             & 72.8             & 56.7             & 50.0             & 72.1             & 90.8             & 55.4             & 61.5 \\
\textbf{EmoLA~\citep{emola}}                       & 57.4             & \textbf{52.4}    & 61.0             & 78.1             & 77.8             & 81.9             & \textbf{89.5}    & 60.5             & 49.3             & 64.9             & 46.0             & 52.4             & 64.2             & 50.5             & 56.9             & \textbf{83.5}    & 55.2             & 43.1             & \underline{80.1} & 91.6             & 60.0             & 65.1 \\
\textbf{FG-Net~\citep{yin2024fg}}                  & -                & -                & -                & -                & -                & -                & -                & -                & -                & -                & -                & -                & 64.3             & -                & -                & -                & -                & -                & -                & -                & -                & 65.4 \\
\textbf{KSRL~\citep{chang2022knowledge}}           & 53.3             & 47.4             & 56.2             & 79.4             & \textbf{80.7}    & 85.1             & 89.0             & 67.4             & 55.9             & 61.9             & 48.5             & 49.0             & 64.5             & 60.4             & 59.2             & 67.5             & 52.7             & 51.5             & 76.1             & 91.3             & 57.7             & 64.5 \\
\textbf{ReCoT~\citep{li2023recot}}                 & 51.5             & 47.8             & 58.9             & 79.2             & \underline{80.2} & 84.9             & 88.4             & 61.5             & 53.3             & 64.6             & \underline{51.8} & 55.4             & 64.8             & 51.3             & 36.2             & 66.8             & 50.1             & 52.4             & 78.8             & \textbf{95.3}    & 69.7             & 62.6 \\
\textbf{MEGraph~\citep{luo2022learning}}           & 52.7             & 44.3             & 60.9             & 79.9             & 80.1             & 85.3             & 89.2             & 69.4             & 55.4             & 64.4             & 49.8             & 51.1             & 65.5             & 52.5             & 45.7             & 76.1             & 51.8             & 46.5             & 76.1             & 92.9             & 57.6             & 62.4 \\
\textbf{CLEF~\citep{zhang2023weakly}}              & 55.8             & 46.8             & \underline{63.3} & 79.5             & 77.6             & 83.6             & 87.8             & 67.3             & 55.2             & 63.5             & \textbf{53.0}    & 57.8             & 65.9             & 64.3             & 61.8             & 68.4             & 49.0             & 52.2             & 72.9             & 89.9             & 57.0             & 64.8 \\
\textbf{MCM~\citep{zhang2024multimodal}}           & 54.4             & 48.5             & 60.6             & 79.1             & 77.0             & 84.0             & 89.1             & 61.7             & \textbf{59.3}    & 64.7             & \textbf{53.0}    & \textbf{60.5}    & 66.0             & 49.6             & 44.1             & 67.2             & \underline{65.5} & 49.0             & \textbf{81.5}    & 85.9             & 71.8             & 64.3 \\
\textbf{MDHRM~\citep{wang2024multi}}               & 54.6             & \underline{49.7} & 61.0             & 79.9             & 79.4             & 85.4             & 88.5             & 67.8             & \textit{56.8}    & 63.2             & 50.9             & 55.4             & 66.1             & 65.4             & 60.2             & 75.2             & 50.2             & 52.4             & 74.3             & \textit{93.7}    & 58.2             & 66.2 \\
\textbf{AUFormer~\citep{yuan2024auformer}}         & -                & -                & -                & -                & -                & -                & -                & -                & -                & -                & -                & -                & 66.2             & -                & -                & -                & -                & -                & -                & -                & -                & 66.4 \\
\bottomrule                                                                                                                                                                                                                                                                                              
\rowcolor{gray!10}                                                                                                                                                                                                                                                                                              
\textbf{CausalAffect (SG)}                         & 63.4             & 47.1             & 56.8             & \textbf{82.1}    & 78.8             & \underline{85.8} & 88.8             & \textit{71.4}    & \underline{57.5} & \textbf{67.9}    & 46.2             & 56.4             & \textit{66.9}$\pm$0.2    & 54.8             & 61.7             & 75.4             & 50.8             & 62.6             & 67.5             & 89.5             & \textit{73.3}    & 67.0$\pm$0.1 \\
\bottomrule                                                                                                                                                                                                                                                                                                                                                                                                                                                                                                                                                                                                     
\rowcolor{gray!10}                                                                                                                                                                                                                                                                                                                                                                                                                                                                
\textbf{CausalAffect (+GFT)}                       & 63.7             & 41.4             & 60.7             & 80.0             & 76.8             & \textit{85.7}    & 88.5             & \textbf{71.5}    & 55.9             & \textit{67.0}    & \textit{51.4}    & 57.1             & 66.6$\pm$0.1             & 56.7             & \textbf{66.1}    & 77.0             & \textbf{68.3}    & \textit{63.7}    & 72.2             & 91.9             & 68.6             & \textit{70.6}$\pm$0.3 \\
\rowcolor{gray!10}                                                                                                                                                                                                                                                                                                                                                                                                                                                                
\textbf{CausalAffect (+EmotioNet)}                 & \textbf{67.1}    & 43.6             & \textbf{66.0}    & 80.1             & 79.1             & 84.8             & \textit{88.9}    & 71.1             & 55.6             & 66.6             & 47.5             & \textit{58.8}    & \textbf{67.4}$\pm$0.3    & \underline{68.1} & 63.2             & \textit{77.6}    & \textit{64.1}    & \textbf{74.0}    & 69.3             & 83.7             & 68.7             & \underline{71.1}$\pm$0.2 \\ 
\bottomrule                                                                                                                                                                                                                                                                                                                                                                                                                                                                                                                                                                                                                                          
\rowcolor{gray!10}                                                                                                                                                                                                                                                                                                                                                                                                                                                                                  
\textbf{CausalAffect (+RAF-DB)}                    & 63.2             & 45.7             & \textit{61.4}    & \textit{81.2}    & 79.1             & 85.1             & 88.5             & 71.0             & 55.3             & \textit{67.0}    & 51.3             & 56.5             & \underline{67.1}$\pm$0.1             & \textit{65.6}    & \textit{65.2}    & 73.4             & 56.5             & 61.6             & 70.5             & 89.2             & \underline{74.4} & 69.5$\pm$0.2 \\
\rowcolor{gray!10}                                                                                                                                                                                                                                                                                                                                                                                                                                                                                                           
\textbf{CausalAffect (+AffectNet)}                 & \textit{64.4}    & 44.6             & 61.2             & \underline{81.3} & \underline{80.2} & 85.5             & 88.8             & \textbf{71.5}    & 52.6             & \underline{67.4} & 45.5             & 56.3             & 66.6$\pm$0.2             & 59.0             & \underline{65.7} & 72.3             & 62.4             & 58.3             & 69.5             & 83.7             & \textbf{74.8}    & 68.2$\pm$0.2 \\
\bottomrule                                                                                                                                                                                                                                                                                                                                                                                                                                                                                                                                                                                                                                          
\rowcolor{gray!10}                                                                                                                                                                                                                                                                                                                                                                                                                                                               
\textbf{CausalAffect (+All)}                       & \underline{65.3} & 42.1             & 58.0             & 81.0             & 78.8             & 85.0             & 87.9             & 71.0             & 56.3             & 66.0             & 48.7             & \underline{60.3} & 66.7$\pm$0.2             & \textbf{72.5}    & 63.7             & \underline{80.0} & 51.9             & \underline{67.1} & 75.0             & 92.6             & 69.0             & \textbf{71.5}$\pm$0.2 \\
\bottomrule                                                                                                                                                                                                                                                                                              
\end{tabular}
\end{adjustbox}
\end{subtable}

\vspace{0.5em} 

\begin{subtable}{\textwidth}
\centering
\begin{minipage}[t]{0.3\textwidth}
\resizebox{\linewidth}{!}{ 
\setlength{\tabcolsep}{20pt}  
\renewcommand{\arraystretch}{1.1}
\begin{tabular}{cc}
\toprule
\textbf{EmotioNet (AU, Image DB)} & \textbf{F1} \\
\midrule
Res50             & 44.0 \\
ME-GraphAU~\citep{luo2022learning}          & 64.9 \\
JAA-Net~\citep{jaa2018}             & 51.8 \\
AUNets~\citep{romero2022multi}              & 64.6 \\
FBNet~\citep{kollias2019face}               & 54.0 \\
CTC~\citep{zhou2023leveraging}                 & 64.4 \\
FUXI~\citep{zhang2023multi}                & \textit{65.4} \\
SITU~\citep{liu2023facial}                & 64.2 \\
\bottomrule
\rowcolor{gray!10}   
\textbf{CausalAffect (SG)  }         & \textbf{66.4}$\pm$0.1 \\
\bottomrule
\rowcolor{gray!10}   
\textbf{CausalAffect (+DISFA)}       & \underline{66.0}$\pm$0.3 \\
\rowcolor{gray!10}   
\textbf{CausalAffect (+BP4D)}        & \textit{65.4}$\pm$0.1 \\
\rowcolor{gray!10}   
\textbf{CausalAffect (+All) }        & 65.0$\pm$0.1 \\
\bottomrule
\end{tabular}}
\end{minipage}
\hfill
\begin{minipage}[t]{0.3\textwidth}
\resizebox{\linewidth}{!}{ 
\setlength{\tabcolsep}{22pt}  
\renewcommand{\arraystretch}{1}
\begin{tabular}{lc}
\toprule
\textbf{GFT (AU, Video DB)} & \textbf{F1} \\
\midrule
EAC-Net~\citep{li2018eac}             & 46.1 \\
TCAE~\citep{li2019self}                & 44.2 \\
CDAU~\citep{ertugrul2020crossing}                & 45.3 \\
ARL~\citep{shao2019facial}                 & 50.1 \\
MoCo~\citep{he2020momentum}                & 52.4 \\
TR~\citep{lu2020self}                  & 54.7 \\
JÂA-Net~\citep{jaa2021}             & 53.7 \\
EmoCo~\citep{sun2021emotion}               & 58.6 \\
EmoLA~\citep{li2024facial}               & \textit{62.1} \\
\bottomrule
\rowcolor{gray!10}   
\textbf{CausalAffect (SG)}           & 61.1$\pm$0.2 \\
\bottomrule
\rowcolor{gray!10}   
\textbf{CausalAffect (+DISFA)}       & \textbf{62.5}$\pm$0.3 \\
\rowcolor{gray!10}   
\textbf{CausalAffect (+BP4D)}        & 60.4$\pm$0.4 \\
\rowcolor{gray!10}   
\textbf{CausalAffect (+All) }        & \underline{62.4}$\pm$0.3 \\
\bottomrule
\end{tabular}}
\end{minipage}
\hfill
\begin{minipage}[t]{0.38\textwidth}
\resizebox{\linewidth}{!}{ 
\setlength{\tabcolsep}{13.5pt}  
\renewcommand{\arraystretch}{1}
\begin{tabular}{lcc}
\toprule
\textbf{AffectNet / RAF-DB (Expr, Image DB)} & \textbf{AffN.} & \textbf{RAF.} \\
\midrule
VTFF~\citep{ma2021facial}               & 61.9 & 81.2 \\
MViT~\citep{li2021mvt}               & 64.6 & 80.4 \\
FBNet~\citep{kollias2021distribution}              & 65.0 & 78.0 \\
DACL~\citep{farzaneh2021facial}               & 65.2 & 80.4 \\
ARM~\citep{shi2021learning}                & 65.2 & 82.8 \\
Ad-Corre~\citep{fard2022ad}           & 63.4 & 79.0 \\
DDAMFN++~\citep{zhang2023dual}           & \underline{67.4} & 84.6 \\
FRA~\citep{gao2024self}                  & 66.1 & 83.4 \\
CAGE~\citep{wagner2024cage}              & 66.4 & 83.2 \\
ExpLLM~\citep{zhang2025lofi}              & 65.9 & - \\  
LOFI~\citep{lan2025expllm}              & 65.7 & \underline{84.7} \\             
\bottomrule
\rowcolor{gray!10}   
\textbf{CausalAffect (+DISFA)}      & 65.6$\pm$0.2 & 83.7$\pm$0.1 \\
\rowcolor{gray!10}   
\textbf{CausalAffect (+BP4D) }      & \textbf{67.7}$\pm$0.1 & \textbf{85.3}$\pm$0.2 \\
\rowcolor{gray!10}   
\textbf{CausalAffect (+All)}        & \textit{66.5}$\pm$0.1 & \textit{84.9}$\pm$0.2 \\
\bottomrule
\end{tabular}}
\end{minipage}
\end{subtable}

\label{tab:all_results}
\end{table*}

\begin{figure}[htbp]
\centering
\includegraphics[width=1\textwidth]{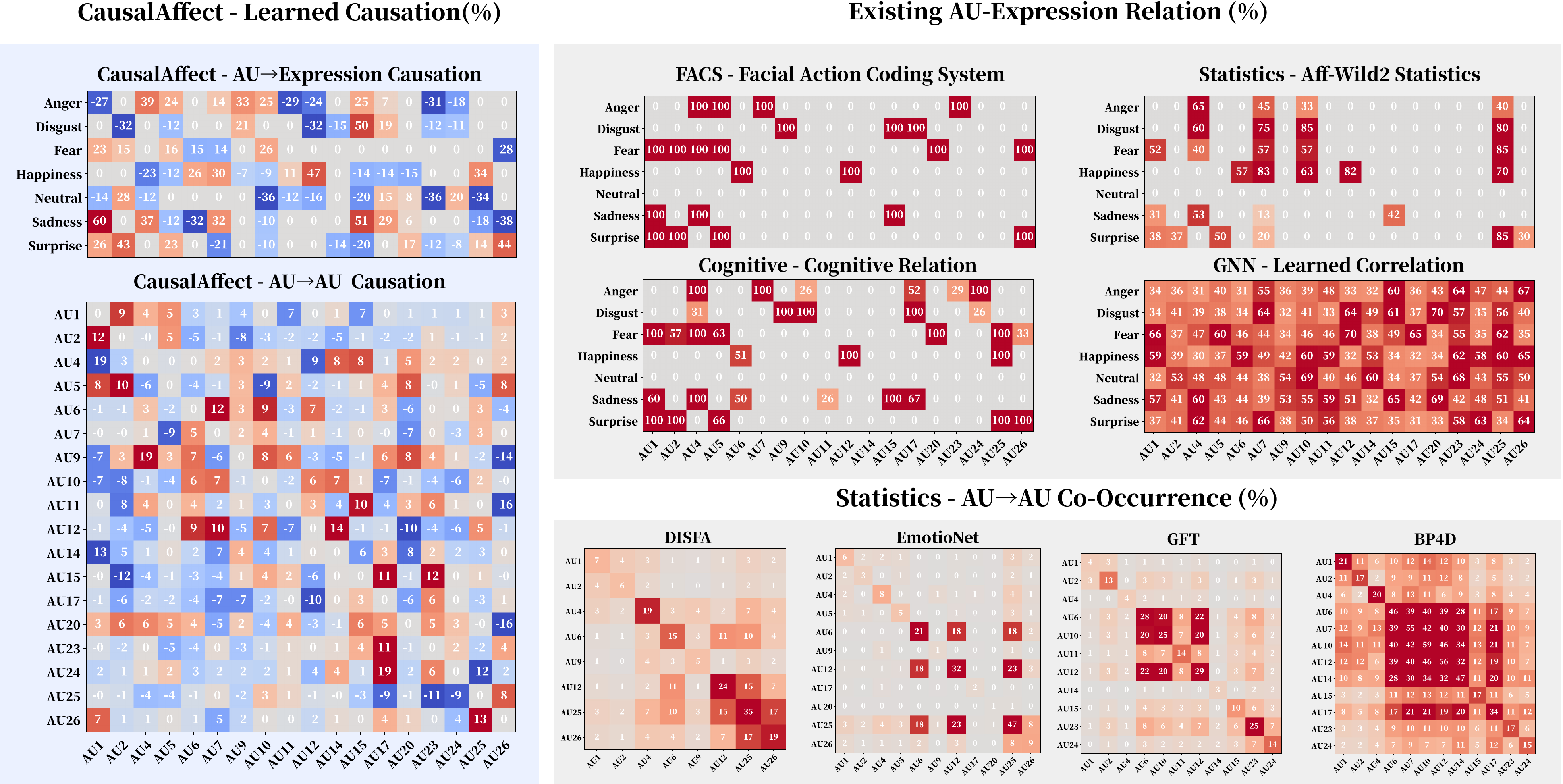}
\caption{
{\footnotesize
Comparison of CausalAffect-learned causal relations (trained on All-DB) with existing approaches. 
\textbf{For AU$\rightarrow$Expression relations}, we compare against FACS~\citep{ekman1978facial}, cognitive priors~\citep{du2014compound}, statistical co-occurrence from Aff-Wild2~\citep{kollias2024distribution}, and GNN-learned correlations~\citep{kipf2016semi}. 
\textbf{For AU$\rightarrow$AU relations}, we compare co-occurrence statistics derived from four commonly used datasets.}
}
\label{fig:au_expr_heatmap}
\end{figure}

\subsection{Comparison of CausalAffect with state-of-the-art (SOTA) }
\label{sec:sota}
\textbf{For AU detection}, CausalAffect surpasses all prior SOTA methods under the single-dataset setting (SG). Incorporating additional AU datasets (e.g., BP4D+EmotioNet, DISFA+GFT) further improves performance by expanding AU categories and enabling transferable causal structures, while auxiliary supervision from expression datasets (e.g., BP4D+RAF-DB, DISFA+AffectNet) brings additional gains.  
\textbf{For expression recognition}, CausalAffect outperforms SOTA methods on AffectNet and RAF-DB without expression-specific encoders—predictions are causally derived from AU features. Richer AU supervision (e.g., BP4D 12 AUs vs. DISFA 8 AUs) provides stronger causal priors, enhancing both accuracy and interpretability (see \textbf{Appendix B} for sensitivity to AU composition).

\subsection{Global Causal Relation Analysis}
\label{sec:global}
\textbf{CausalAffect vs. Existing Relations.}  
As shown in Figure~\ref{fig:au_expr_heatmap}, prior AU$\rightarrow$Expression relations (FACS, cognitive, statistical, or GNN-based) rely only on hard-coded excitatory links or yield dense entangled patterns. In contrast, CausalAffect learns \textbf{polarity-aware, sparse graphs} that capture both excitatory and inhibitory pathways. For AU$\rightarrow$AU, existing methods reduce to symmetric co-occurrence within single datasets, missing the \textbf{asymmetric interactions} observed in psychology~\citep{rinn1984neuropsychology},whereas CausalAffect integrates non-overlapping datasets to construct asymmetric dependencies in a psychologically consistent manner (see \textbf{Appendix C} for detailed analysis).

\begin{table*}[h]
\centering
\caption{
{\small
Comparative AU$\rightarrow$Expression relations from prior literature and our global causal graph. \textbf{Bold} entries denote AUs reported in the literature and also identified as positive by CausalAffect.}}
\renewcommand{\arraystretch}{1}
\resizebox{\linewidth}{!}{ 
\setlength{\tabcolsep}{3pt}  
\renewcommand{\arraystretch}{1}
\begin{tabular}{lcccccc}
\toprule
 & \textbf{\citep{ekman1976measuring}} & \textbf{\citep{lucey2010extended}} & \textbf{\citep{karthick2013survey}} & \textbf{\citep{du2014compound}} & \textbf{\citep{clark2020facial}} & \textbf{Ours (Global Causal Graph)} \\
\midrule
\textbf{Sadness}   & \textbf{1, 4, 15}       & \textbf{1, 4}, 11, \textbf{15, 17} & \textbf{1, 4,  15, 17}    & \textbf{1, 4}, 6, 11, \textbf{15, 17} & \textbf{1, 4, 15, 17} & \textbf{1, 4}, 7, \textbf{15, 17} \\
\textbf{Surprise}  & \textbf{1, 2, 5, 26}    & \textbf{1, 2, 5, 25}, 27  & \textbf{1, 2, 5, 26}, 27    & \textbf{1, 2, 5, 25, 26}     & \textbf{1, 2, 5, 26}, 27 & \textbf{1, 2, 5, 20, 25, 26} \\
\textbf{Happy}     & \textbf{6, 12}          & \textbf{6, 12, 25}        & \textbf{6, 12, 25}          & \textbf{6, 12, 25}           & \textbf{6, 12} & \textbf{6}, 7, 11, \textbf{12, 25} \\
\textbf{Disgust}   & \textbf{9, 15, 17}      & \textbf{9}, 10, \textbf{15, 17}    & \textbf{9, 17}              & 4, \textbf{9}, 10, \textbf{17}, 24    & \textbf{9}, 10, \textbf{17} & \textbf{9, 15, 17} \\
\textbf{Fear}      & \textbf{1, 2}, 4, \textbf{5}, 20, 26 & \textbf{1, 2}, 4, \textbf{5}   & \textbf{1}, 4, \textbf{5}, 7         & \textbf{1, 2}, 4, \textbf{5}, 20, 25, 26 & \textbf{1, 2}, 4, \textbf{5}, 20, 25 & \textbf{1, 2, 5}, 10 \\
\textbf{Anger}     & \textbf{4, 5, 7}, 23    & \textbf{4, 5, 9, 10, 15, 17}, 23, 24 & \textbf{4, 5, 7}, 23, 24 & \textbf{4, 7, 10, 17}, 23, 24 & \textbf{4, 5, 7, 10, 17}, 23, 25 & \textbf{4, 5, 7, 9, 10, 15, 17} \\
\bottomrule
\end{tabular}
}

\label{tab:lit_compare}
\end{table*}
\paragraph{Psychological Plausibility.}  
To assess human alignment, we compare CausalAffect’s discovered relations with well-established findings (Table~\ref{tab:lit_compare}). The learned graphs exhibit strong consistency with canonical psychological FACS~\citep{ekman1976measuring}, cognitive studies~\citep{du2014compound}, and empirical mappings~\citep{lucey2010extended,clark2020facial}. Beyond recovering these canonical pathways, CausalAffect reveals \textbf{novel and literature-supported relations}, such as AU7$\rightarrow$Sadness~\citep{miller2022observers} and AU10$\rightarrow$Fear~\citep{blasberg2023you}, as well as \textbf{inhibitory effects} (e.g., AU6$\dashv$Sadness, AU26$\dashv$Sadness) highlighting the role of suppression in reliable emotion inference. These findings demonstrate that CausalAffect not only expands structural expressiveness beyond prior methods but also yields psychologically grounded causal relations (More evidence in \textbf{Appendix C.1}).
%

%
%
\subsection{Case Study: Sapmle-Adaptive Causal Relation Analysis}
\label{sec:sa}
\begin{figure}[htbp]
\centering
\includegraphics[width=1\textwidth]{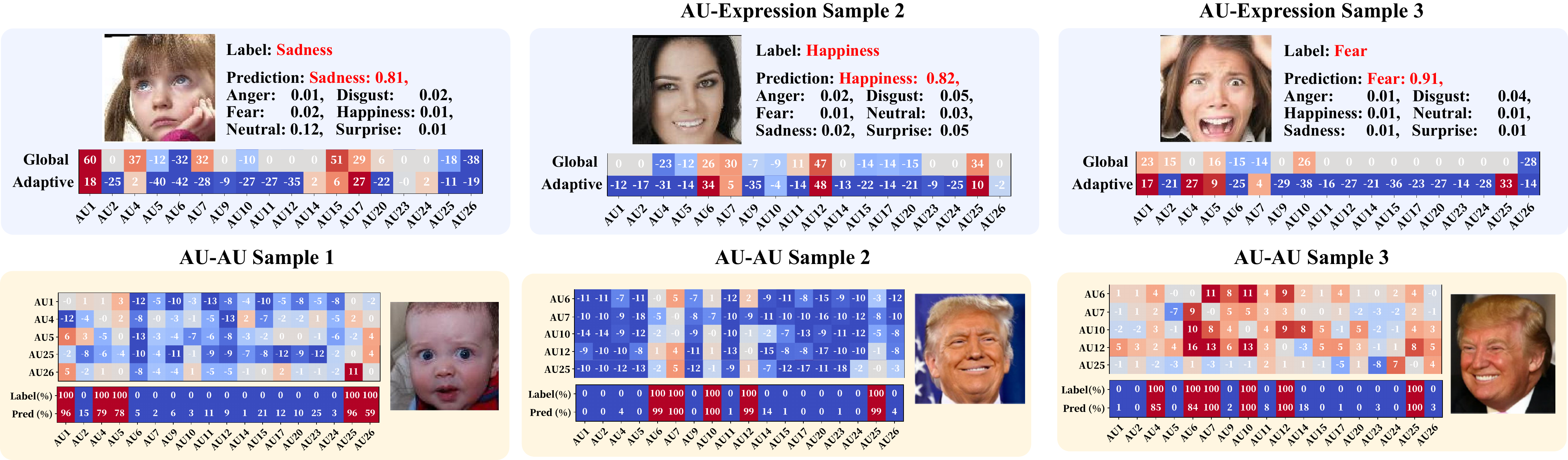}
\caption{
{\small
Sample-adaptive causal relations inferred for individual cases. Examples are shown for AU$\rightarrow$Expression (top, blue) and AU$\rightarrow$AU (bottom, yellow), illustrating how CausalAffect adapts to context-specific structures. For clarity, only subgraphs involving the predicted expression or active AUs are visualized.}
}

\label{fig:case}
\end{figure}

\textbf{For AU$\rightarrow$Expression}, the sample-adaptive graph (Figure~\ref{fig:case}, upper blue) demonstrates the ability to remain consistent with the global structure while \textit{capturing instance-specific adaptations}. In \textit{Sample 2}, the SAC-Graph highlights AU6, AU12, and AU25 as dominant contributors, closely aligning with the Global Graph. Meanwhile, it shows strong \textbf{\textit{adaptability}} to contextual cues: in \textit{Sample 1}, AU17 (chin raiser) is prioritized for its visual salience, whereas the Global Graph emphasizes the inactive AU4 (brow lowerer); in \textit{Sample 3}, AU4 and AU25 are elevated for their visible activation, while globally emphasized but inactive AUs such as AU10 and AU2 are downweighted. 

\textbf{For AU$\rightarrow$AU} (Figure~\ref{fig:case}, lower yellow), \textit{Sample 1} remains consistent with the Global Graph, where AU26 is driven by AU25 and AU5 by AU1/2/26. Beyond these canonical links, CausalAffect uncovers that many \textbf{non-root AUs} emerge not through direct excitation but through the absence of their suppressors, highlighting the role of \textbf{inhibitory dependencies}. For example, in \textit{Sample 1} AU4 (brow lowerer) is largely inferred via suppression by AU1 (inner brow raiser), consistent with their antagonistic muscular relation in psychology~\citep{karmann2015role,cattaneo2007inhibitory}. Moreover, \textbf{psychological plausibility} is evident when comparing samples with similar active AUs but divergent structures~\citep{ekman1990duchenne,surakka1998facial}: \textit{Sample 2 (Social Smile)} is dominated by inhibitory links, indicative of strained or socially modulated expressions, whereas \textit{Sample 3 (Duchenne Smile)} exhibits a coherent feedforward graph where AUs mutually reinforce one another, consistent with spontaneous, genuine smiles. These results demonstrate CausalAffect’s ability to model context-dependent dynamics, aligning with theories of expressive modulation. (More analysis in \textbf{Appendix D})

\subsection{Ablation Study}
\label{sec:ablation}

To better understand the contributions of individual modules and their interactions, we conduct an extensive ablation study on CausalAffect (+All) using 15 model variants (Rows 1--15 in Table~\ref{tab:ablation}). The ablation study highlights the complementary roles of different modules. The GC captures stable population-level dependencies, while the SAC Graph introduces context-adaptive flexibility. Disentanglement provides psychologically meaningful features. Counterfactual further prunes spurious associations but only becomes effective when disentangled representations are available. Finally, the DAG constraint ensures structural clarity and interpretability.  (More analysis in \textbf{Appendix A})

\begin{table*}[h]
\centering
\renewcommand{\arraystretch}{1.2}
\normalsize
\resizebox{\linewidth}{!}{ 
\setlength{\tabcolsep}{10pt}  
\renewcommand{\arraystretch}{1}
\begin{tabular}{c l l c c c c c c}
\toprule
\textbf{Idx} & \textbf{Model Variant} & \textbf{w/o} & \textbf{AffectNet} & \textbf{RAF-DB} & \textbf{DISFA} & \textbf{BP4D} & \textbf{GFT} & \textbf{EmotioNet} \\
\midrule
1  & Backbone                     & GC + SAC + CF        & 58.9 & 70.0 & 53.0 & 57.2 & 57.9 & 47.5 \\
2  & Backbone + Dis               & GC + SAC + CF        & 57.1 & 69.3 & 54.2 & 55.5 & 57.8 & 59.2 \\
\midrule
3  & Backbone + GC                & Dis + CF + SAC       & 62.3 & 80.2 & 62.4 & 61.0 & 60.9 & 61.7 \\
4  & Backbone + GC + Dis          & CF + SAC             & 61.9 & 78.2 & 61.1 & 59.8 & 60.5 & 61.4 \\
5  & Backbone + GC + CF           & Dis + SAC            & 62.5 & 79.8 & 61.5 & 62.1 & 61.7 & 62.3 \\
6  & Backbone + GC + Dis + CF     & SAC                  & 64.4 & 83.3 & 65.8 & 66.6 & 61.0 & 63.6 \\
\midrule
7  & Backbone + SAC                & Dis + CF + GC       & 62.0 & 78.1 & 60.5 & 61.1 & 58.9 & 60.7 \\
8  & Backbone + SAC + Dis          & CF + GC             & 60.7 & 77.9 & 60.1 & 59.5 & 57.9 & 59.2 \\
9  & Backbone + SAC + CF           & Dis + GC            & 61.3 & 77.5 & 60.6 & 60.7 & 57.5 & 60.2 \\
10 & Backbone + SAC + Dis + CF     & GC                  & 62.7 & 78.5 & 61.7 & 62.4 & 59.1 & 61.2 \\
\midrule
11 & Backbone + GC + SAC           & Dis + CF            & 63.1 & 82.9 & 64.1 & 61.5 & 60.4 & 62.1 \\
12 & Backbone + GC + SAC + Dis     & CF                  & 62.6 & 81.4 & 66.4 & 61.3 & 60.1 & 62.0 \\
13 & Backbone + GC + SAC + CF      & Dis                 & 64.3 & 83.4 & 62.9 & 62.8 & \textbf{62.6} & 63.4 \\
\midrule
14 & CausalAffect (w/o DAG)               & w/o DAG             & 65.5 & 84.6 & 71.3 & 64.4 & 62.1 & 64.7 \\
\midrule
\rowcolor{gray!10}  
15 & \textbf{CausalAffect (GC + SAC + Dis + CF)} & /            & \textbf{66.5} & \textbf{84.9} & \textbf{71.5} & \textbf{66.7} & 62.4 & \textbf{65.0} \\
\bottomrule
\end{tabular}}
\caption{
{\small
Ablation Study on CausalAffect (+All Setting), exploring the effect of Global Causal Graph (GC), Sample-Adaptive Causal Graph (SAC), Counterfactual Intervention (CF), and AU Disentanglement (Dis).
}}
\label{tab:ablation}
\end{table*}

\section*{Conclusion}
We presented CausalAffect, a weakly supervised framework for discovering \emph{polarity-aware}, \emph{directed}, and \emph{sample-adaptive} causal relations among facial AUs and between AUs and expressions. By jointly constructing a population-level global graph and a sample-adaptive graph, and by enforcing node-level counterfactual interventions, our method recovers psychologically plausible structures that align with established cognitive studies while also uncovering inhibitory pathways and novel relations. Across six benchmarks, CausalAffect advances the state of the art in both AU detection and expression recognition, yielding human-aligned and interpretable causal graphs.

\bibliography{iclr2026_conference}
\bibliographystyle{iclr2026_conference}

\end{document}

%% file: math_commands.tex
\usepackage{amsmath,amsfonts,bm}

\def\eqref#1{equation~\ref{#1}}

\def\1{\bm{1}}

\DeclareMathAlphabet{\mathsfit}{\encodingdefault}{\sfdefault}{m}{sl}
\SetMathAlphabet{\mathsfit}{bold}{\encodingdefault}{\sfdefault}{bx}{n}

